# Multi-class Regret Detection in Hindi Devanagari Script

**Renuka Sharma[a], Sushama Nagpal[a], Sangeeta Sabharwal[a], Sabur Butt[b]**

[a]Department of Computer Science and Engineering, Netaji Subhas University of Technology
[b]Institute for the Future of Education, Tecnológico de Monterrey, 64849, Mexico
{renuka.sharma.phd22, sushma.nagpal, ssab}@nsut.ac.in, saburb@tec.mx

## Abstract

The number of Hindi speakers on social media has increased dramatically in recent years. Regret is a common emotional experience in our everyday life. Many speakers on social media, share their regretful experiences and opinions regularly. It might cause a re-evaluation of one's choices and a desire to make a different option if given the chance. As a result, knowing the source of regret is critical for investigating its impact on behavior and decision-making. This study focuses on regret and how it is expressed, specifically in Hindi, on various social media platforms. In our study, we present a novel dataset from three different sources, where each sentence has been manually classified into one of three classes "Regret by action", "Regret by inaction", and "No regret". Next, we use this dataset to investigate the linguistic expressions of regret in Hindi text and also identify the textual domains that are most frequently associated with regret. Our findings indicate that individuals on social media platforms frequently express regret for both past inactions and actions, particularly within the domain of interpersonal relationships. We use a pre-trained BERT model to generate word embeddings for the Hindi dataset and also compare deep learning models with conventional machine learning models in order to demonstrate accuracy. Our results show that BERT embedding with CNN consistently surpassed other models. This described the effectiveness of BERT for conveying the context and meaning of words in the regret domain.

## 1 Introduction

Social media plays a distinct role in modern societies, enabling individuals to share, tweet, and post virtually anything (Shulman and Simo, 2021). Regret is a feeling of sadness about something sad or wrong or about a mistake that you have made, and a wish that it could have been different and better (Matsumoto, 2009). It is a negative, cognitive emotion experienced when we get a completely different response from our decisions that do not match the expected result (Saffrey et al., 2008). It is frequently associated with feelings of remorse, self-blame, and disappointment. (Budjanovcanin and Woodrow, 2022; Michenaud and Solnik, 2008). It is a psychological emotion that can influence our reactions to decisions, and its outcome can also depend on prior decision outcomes. However, it can also motivate a person to think about how this incident could have occurred and what could be done to stop it from happening again. (Zeelenberg et al., 2002).

Regret is frequently the result of past actions or decisions a person makes or the actions and decisions they neglect to take (Gilovich and Medvec, 1995). According to (Zeelenberg et al., 1998), the intensity and duration of regret may vary based on its origin, with some suggesting that regret stemming from an action can result in immediate pain but may not have enduring consequences. while the regret associated with not taking action might be more prolonged and troublesome. Regret emphasizes its role in decision-making (Bleichrodt et al., 2010), particularly in uncertain situations, where people care about the outcome and consider potential outcomes if they had made a different decision. However, (Diecidue and Somasundaram, 2017) implies that regret might cause a re-evaluation of one's choices and a desire to make a different option if given the chance. As a result, knowing the source of regret is critical for investigating its impact on behavior and decision-making. Researchers have revealed, that the impact of regret on behavior and decisions varies across different life domains, encompassing areas such as education, career, romance, parenting, self-improvement, and leisure activities among the American population. (Roese and Summerville, 2005; Balouchzahi et al., 2023).

Many people like to express their feelings of regret and disappointment in the language they are more comfortable with. The majority of people in India speak Hindi as their native language. According to (Kulkarni and Rodd, 2021), approximately 500 million people globally use the Hindi language. Hindi is one of the 22 official languages acknowledged by the Indian Constitution. Approximately 44% of the Indian population speaks Hindi or its various dialects, making it the most widely spoken language in India (Jain et al., 2020). India is a diverse nation with multiple scripts and languages, and one of the most prevalent scripts is Devanagari, which is employed for writing Hindi (Narang et al., 2020). The aim of this study is to provide a computational way to understand regret in natural language processing (NLP) applications. It seeks to comprehend the role of regret in Hindi text and multimodal human emotions and decision-making processes. The objective is to create more effective methods for dealing with regret as well as to increase the accuracy and efficacy of NLP applications in many sectors.

To our understanding, this is the first work to examine the regret detection challenge in the Hindi dataset. Performing regret detection in Hindi text poses several significant challenges. One major hurdle is the flexible word order in Hindi compared to the fixed Subject-Verb-Object (SVO) structure in English. This variability in sentence structure can affect the polarity of the text, making linguistic analysis in Hindi more complex. Additionally, the morphological richness of Hindi results in a single root word generating various related words, complicating emotion analysis. Variations in spelling, due to the phonetic nature of the language and diverse dialects, further add to the complexity. These challenges require a deep linguistic understanding and specialized handling to perform accurate regret detection in Hindi.

The main intent of this paper is to view and analyze the foremost annotated dataset from three separate sources on regret detection in Hindi literature. The dataset was created by manual scraping, using an online survey form (Questionaries' method), and YouTube comment extractor methods in the timeline of 01-04-2023 to 30-08-2023. Inspired by (Balouchzahi et al., 2023), we approximated the problem of regret identification by first dividing the dataset into three categories: "Action" (regret by action), "Inaction" (regret by inaction), and "No regret" and second, identifying the domain of the text (for more detail refer to 3.2 Annotation guideline, 3.3 Annotation Details). This research indicates that individuals across various platforms tend to express more regret about their past actions, especially in the context of relationships. Additionally, the study conducted several baseline and state-of-the-art experiments to benchmark the dataset. Our research indicated that BERT (Bidirectional Encoder Representations from Transformers) is highly effective in capturing the meaning and context of words related to regret in Hindi text. Deep learning models utilizing BERT word embedding outperformed other models in all of our experiments.

In summary, the major contributions of this research are:

- The development of the first multi-source regret dataset in Hindi,
- An in-depth analysis of the origins of different types of regret,
- A set of deep learning and machine learning models to benchmark the dataset,
- A comprehensive analysis of the shortcomings of our best-performing model and an analysis of the specific errors found in each class within the dataset.

The rest of the work is organized as follows. Section 2 explains the work related to regret detection, identification, and techniques used. Dataset development, characteristics, limitations, and annotation procedure are mentioned in section 3. The section 4 discusses the detailed methodology for the paper, while section 5 gives the performance analysis and results. Lastly, section 7 and 8 conclude the paper and highlight potential future work.

## 2 Literature Review

NLP applications can benefit from understanding the emotional aspect of regret in various scenarios. Emotional context- regret is accompanied by other feelings, such as (sadness, anger, guilt, and disappointment) depending on the circumstances that lead to the regretful situation. So here, sentiment analysis models can better understand people's fine-grain emotional state and can produce a better opinion summary. Industrial applications like chatbots and virtual assistants can understand user regret and respond empathetically (Balouchzahi et al., 2023). This understanding can improve content moderation, detect sexism, fake news, and offensive language, and improve customer service by analyzing feedback and identifying issues. Recognizing regret in the NLP context can bridge the language-to-actionable insights gap by recognizing and processing regret in text data. Regret has rarely been examined computationally, despite being frequently experienced and significantly impacting human life and behavior

### 2.1 Datasets available for Regret

Simo and Kreutzer (2022) gathered a dataset of 5,420,100 regrettable texts from a variety of news websites and social media platforms, including Facebook, Twitter, 4chan, Reddit, and YouTube channels. Eight regret-related labels were proposed by them for data annotation, based on (Wang et al., 2011) empirical investigation of regrets on Facebook. They also proposed regret-specific pre-trained embedding models. (Shulman and Simo, 2021; Simo and Kreutzer, 2022) such as offensive language, vulgarity, hate speech, and private or important information that might harm a user's reputation and life. This study suggests that regret has been examined from a privacy-related perspective, not an emotion.

Balouchzahi et al. (2023) shared their research on detecting regret and identifying domains in English text from Reddit posts. They collected unlabelled datasets from subreddits: "regret", "regretful parents", and "confession", which were compiled from January 1, 2000 to September 10, 2022. They annotated the dataset by classifying posts into ternary classes: "Action" (regret caused by action), "Inaction" (regret caused by inaction), and "No regret". Similar way, they identified the domain of the text into five predefined categories: "Education", "Health", "Career and Finance", "Romance and Relationships", and "Other domains". In order to evaluate the dataset, they used traditional machine learning and deep learning models. They discovered that GloVe word embedding stacked with Neural Networks (NNs) performed best with averaged macro f1-scores of 0.72% for regret detection and 0.63% for domain identification subtasks. The author faced difficulty in distinguishing between regret stemming from actions and regret arising from inaction due to their similar expression and context-dependent nature in text data. They also address the imbalanced label distribution in the dataset for future analysis. The existing dataset related to regret is presented in Table 1.

### 2.2 Methodologies used for Regret

In the past few studies that have been done (Shulman and Simo, 2021) have shown a growing interest in regrettable posts that could potentially damage a user's life and reputation. In this paper, the author has explored regrettable posts within the realm of social media analysis using NLP with a focus on their identification and analysis. The objective of these investigations was to identify potentially regrettable disclosures, including hate speech, profanity, offensive language, and private or sensitive data, and reduce the obstacles to the development of efficient automated systems for regret-related post-detection.

Balouchzahi et al. (2023) proposed a computational approach to study regret in English texts from Reddit posts, aiming to understand its role in human emotions and decision-making. To evaluate the dataset, they used conventional machine learning and deep learning models named Support Vector Machine (SVM) with Radial Basis Function and linear kernels, Random Forests (RFC), Logistic Regression (LR), XGBoost, AdaBoost, and CNN (Convolutional neural network), Bi-LSTM. They used two sets of feature representations, based on the type of approach. For each of the traditional ML approaches, they ran two sets of experiments, one using n-gram and then using TF-IDF feature vectors. The same process was applied for Neu-ral Network methods, but now using FastText and Glove embeddings as features. GloVe-built Neural Networks, achieving state-of-the-art results with averaged macro f1-scores of 0.72% in regret detection and 0.63% for domain identification subtasks. Sidorov et al. (2023) analyzes various transformer models' performance in regret and hope speech detection on two new datasets. The Roberta-based model performed well with the highest averaged macro f1-score of 0.83%. The model based on Bert had the highest f1-score, 0.72%. The study also highlights how crucial it is to take pre-training, transformer architectures, and context into account for better performance.

| Year | Authors | Classes | Language | Dataset Size | Summary |
|---|---|---|---|---|---|
| 2023 | (Balouchzahi et al., 2023) | No Regret,<br>Regret by action,<br>Regret by inaction, | English | Reddit post:-<br>3425 | Using natural language processing techniques, the researcher presents the problem of multiclass regret detection and domain identification. Using a proposed dataset, they analyze the work's feasibility in English dataset. |
| 2022 | (Simo and Kreutzer, 2022) | Personal & Family,<br>Misogyny,<br>Profanity & Obscenity,<br>Politics,<br>Religion,<br>Sex,<br>Work & Company,<br>Alcohol & Drugs | English<br>,German | Dataset from various social platform:-5,420,100 | This research sought to detect potentially regretful disclosures on social media platforms.. |

Table 1: Dataset related to Regret

The works mentioned suggest that there has been limited research conducted on regret as an independent emotion. Instead, the focus has primarily been on examining it from the angle of privacy concerns, particularly in terms of comprehending how sensitive content spreads on social media networks and is limited to the English language.

### 2.3 Methodologies used for Fine-grain ed Emotions analysis in Hindi

Emotions are a necessary and important component of our existence (Kumar et al., 2023). It is a subdomain of sentiment analysis aiming to extract more detailed emotions, such as pleasure, anger, sadness, etc., from human languages, rather than providing broad and generalized polarity assignments. Emotions can be single-dimen sional or multidimensional. Understanding human behavior deeply requires investigating emotional data, such as actions, words, and actions that often reflect our actions and feelings. Shome (2021) presented a work of emotion recognition on a novel multilevel Hindi text dataset. The author translated the GoEmotions English dataset to the Hindi dataset involves 58,000 text samples annotated with 28 emotion labels. To evaluate the Hindi dataset, finetuned a Multilingual BERT-based transformer model was used. The model achieves outstanding performance with an overall ROC-AUC score of 0.92. The author discovered that while the model performs well on small paragraphs or phrases, its performance degrades as the amount of text rises.

In (Dhar et al., 2022), machine learning methods have been proposed to identify emotions in Hindi song lyrics. The study collected Hindi song lyrics from Bollywood songs and recognized and classified the emotions arising from the lyrical text of Hindi songs. The writer extracted five distinct feature variations from the lyrics of a specific song. Feature vectors were represented using TF-IDF and Doc2vec. The results showed that the performance of the SVM classifier surpassed the highest accuracy of 66.7%. (Kumar et al., 2023), presented a multilingual BERT transformer model for recognizing emotions in Hindi text. In this study, authors used the unlabelled BHAAV dataset which consists of 20,304 sentences. They manually classified the dataset into five emotional categories Anger, Suspense, Joy, Sadness, and Neutral. To predict Hindi text emotions, the study compares machine learning and deep learning models using word embedding. The mBERT model showed outstanding results, with a balanced accuracy of 93.88%, recall of 93.44%, accuracy of 99.55%, and precision of 94.39% on training data, while loss of 0.3073%, balanced accuracy of 91.84%, recall of 91.74%, auc 98.46% and precision of 92.01% on the testing dataset. Ghosh et al. (2023) created a transformer-based multitask framework (XLMR) to improve the effectiveness of transfer learning techniques by utilizing task-specific data to refine the framework for sentiment and emotion recognition. They also discussed the challenges of code-mixed data in a multitasking environment with emotion detection as an auxiliary task and sentiment detection as a primary task. They manually labeled 20,000 instances of Hinglish in the SentiMix dataset. The proposed multi-task model outperformed the best-performing baseline by 1.31% and the state-of-the-art system by 1.93%, with 71.61% accuracy on sentiment and 63.47% on emotion. The writer encountered a problem during the emotion detection phase since their code-mixed dataset was not balanced regarding the distribution of emotion classes, affecting the accuracy of emotion classification. In a study conducted by (Sasidhar et al., 2020), this research concentrates on detecting and classifying emotions in Hindi-English code-mixed texts. They gathered a dataset comprising 12,000 code-mixed texts and annotated them with emotions such as Happiness, Sadness, and Anger. The researcher used a pre-trained bilingual model for generating feature vectors and applied deep neural networks, specifically CNN-BiLSTM, as the classification models. The outcomes indicated that CNN-BiLSTM attained a classification accuracy of 83.21%. Jain and Kashyap (2023) recently, developed a model to assess the collective public sentiment on Twitter regarding the COVID-19 pandemic using a code-mixed dataset. The model included a sentiment analysis task. A hybrid of CNN and long short-term memory (LSTM) model pair was employed to classify sentiments into positive, negative, and neutral categories. The authors used the meta heuristic-based Grey wolf optimization technique to extract the optimal features from a pre-processed dataset. They compared the performance of the suggested model with other machine learning techniques (RF, Decision Tree, K-NN, Naive Bayes, SVM, CNN, LSTM, LSTM–CNN, and CNN–LSTM). The findings indicate that the proposed ensemble hybrid model achieves impressive results with a classification accuracy of 95.54%, precision of 91.44%, recall of 89.63%, and an F-score of 90.87%.

We observed that machine learning and deep learning models especially associated with different feature extraction and word embedding have given better results on similar tasks. So, directed by (Balouchzahi et al., 2023) (Sidorov et al., 2023) (Shome, 2021) (Kumar et al., 2023) (Pandey and Singh, 2023), we used various forms of word embeddings tf-idf, bi-gram, and BERT as our word feature vectors and deep learning networks such as CNN, LSTM, Bi-LSTM, and BERT.

## 3 Dataset

### 3.1 Statistics

For this research, we present a novel dataset for regret origin and its identification in Hindi posts, and comments from three sources Quora, YouTube, and an online survey form (Questionaries method). We collected 1457 samples of data from the online survey form, 842 data samples from Quora by manual extracting, and 12,000 from YouTube using the comments exporter for YouTube. We manually selected 2,062 Hindi samples from YouTube samples containing Hindi regret and merged them with the samples from the other two resources, resulting in 4,074 unlabelled posts.

In Table 3, we present the statistics related to our dataset. Our analysis indicates that users are prone to communicate feelings of regret, whether stemming from their actions or inaction. The statistics reveal an unequal distribution of labels in the no-regret subtask, which could impact the effectiveness of learning models. To rectify this, we addressed the imbalance by introducing 1,814 unlabeled data points from the BHAAV dataset. It is the largest text corpus used for emotion analysis from Hindi stories. The corpus has 20,304 sentences that were gathered from 230 distinct short stories covering 18 different genres, including inspirational and mystery. We also annotated the class and domain of data with the help of an annotator (for more detail refer to 3.2 Annotation guideline, 3.3 Annotation Details). A domains subtask analysis, as presented in Table 4, highlights that users are less prone to express regret regarding decisions related to their health and other life aspects, but are more inclined to express regret in the context of relationships. Meanwhile, Table 5 presents the post statistics in the training and testing datasets after normalization. The complete dataset comprises a vocabulary of 14,050 words, with an average post length of 3.56 words and an average sentence length of 42.60.

| Text | Regret Classes | Domains Identification |
|---|---|---|
| मैं अपनी जिंदगी में जो बनना चाहती थी वो नहीं बन पाई क्योंकि मैंने अपनी पढ़ाई को कभी गंभीरता से नहीं लिया। अगर मैंने मेहनत करके पढ़ाई की होती तो आज मैं अपनी महत्वाकांक्षाओं को पूरा कर पाती।<br><br>Translated in English: I couldn't become what I wanted to be in my life because I never took my studies seriously. If I had worked hard and studied, I could have fulfilled my ambitions today. | Regret by Inaction | Education |
| मुझे मेरे फैसले का खेद है। मुझे बिना जानकारी के शेयर बाजार में पैसे लगाने का अफसोस है। मेरी पत्नी ने भी कहा था कि दूसरों के चक्कर में आकर मत करो,पहले खुद से जानकारी लो। मैंने तब उनकी बात नहीं सुनी और काफी नुकसान हुआ।<br><br>Translated in English: I regret my decisions. I feel sorry for investing in the stock market without proper knowledge. My wife also advised me not to get involved without understanding first, but I didn't listen to her, and I suffered significant losses. | Regret by Action | Career and Finance |
| 19 साल की उम्र में भाग के शादी करना, वो भी इंटरकास्ट लव मैरिज, इस पर मुझे बहुत पछतावा है। मेरे घरवाले मेरे साथ कोई बात नहीं करते। उनके लिए मैं मर चुकी हूं। घर में कुछ भी होता है तो वो लोग मुझे नहीं बुलाते। मुझे इस तरह की शादी नहीं करनी चाहिए थी। आज मैं अकेली हूं अपने परिवार के बिना।<br><br>Translated in English: I deeply regret getting married at the age of 19, especially in an intercaste love marriage. My family doesn't talk to me anymore. For them, I am as good as dead. They don't invite me for any family events. I should not have gone through with this kind of marriage. Today, I am alone without my family. | Regret by Action | Romance and Relationships |
| मैंने आगे के अध्ययन का समय नहीं निकाला, काश कि मैंने कर लिया होता। दुर्भाग्यवश, बहुत लंबे समय तक मुझे नहीं पता था कि मैं कौन सा करियर चाहता हूँ, इसलिए मुझे यह नहीं पता था कि मुझे क्या पढ़ना चाहिए। और जब मुझे पता चला, तो मेरे पास बहुत सारी जिम्मेदारियां थीं।<br><br>Translated in English: I didn't dedicate enough time to further studies; I wish I had. Unfortunately, for a long time, I didn't know what career path I wanted to pursue, so I didn't know what to study. And when I finally figured it out, I had many responsibilities to handle. | Regret by Inaction | Education |
| मुझे लगता है कि हम अपने किशोरावस्था में बहुत सारी गलतियाँ करते हैं जिन पर हम बाद में पछताते हैं। कभी–कभी हम गलती करने के बाद सोचते हैं, काश किसी ने पहले ही मुझे बता दिया होता कि यह गलत है।<br>Translated in English: I think we make a lot of mistakes in our teens that we later regret. Sometimes after making a mistake, we think, I wish someone had told me earlier that it was wrong. | No Regret | Others |
| मेरे ताऊजी की अचानक मृत्यु होना और उन्हें अंतिम बार नहीं देख पाना परीक्षाओं की वजह से।<br>Translated in English: My uncle's sudden death and not being able to see him for the last time because of exams. | Regret by Inaction | Romance and Relationships |
| नहीं... कभी नहीं, मेरे जीवन में मुझे कोई पश्चाताप नहीं है। मैंने अपने अच्छे दोस्तों के साथ अपने जीवन का पूरा आनंद उठाया, जिन्हें मैं 'लंगोटिया यार' कहता हूँ। और मुझे उनकी बहुत याद आती है और मेरे एक दोस्त अजमेर अब इस दुनिया में नहीं है।<br>Translated in English: No... never, I have no regrets in my life. I enjoyed my life to the fullest with my good friends, whom I call 'Langotiya Yaar'. And I miss him a lot and one of my friends Ajmer is no more in this world. | No Regret | Romance and Relationships |
| हम अपने शारीरिक स्वास्थ्य को ज्यादा समय नहीं दे पाते हैं। जिसके कारण हमें कमर में दर्द रहता है। समय की कमी में हम अपना ध्यान भी नहीं दे पाते। हालांकि, मेरी उम्र अभी 21 साल है।<br><br>Translation in English: We are not able to give much time to our physical health. Due to this, we have back pain. Due to lack of time, we are not able to even concentrate. However, I am still 21 years old. | Regret by Inaction | Health |

Table 2: Sample texts from the dataset.

| Subclass | Different Sources of Dataset | | | | Count |
|---|---|---|---|---|---|
| | Online Survey Data | Quora Post | Youtube Comments | BHAAV Dataset | |
| Regret by Action | 720 | 454 | 948 | - | 2122 |
| Regret by Inaction | 486 | 351 | 1018 | - | 1855 |
| No Regret | 251 | 37 | 96 | 1814 | 2198 |
| Total | 1457 | 842 | 2062 | 1814 | **6175** |

Table 3: Statistics of the dataset and its sources

### 3.2 Annotation Guideline

We gave the annotators detailed instructions and example instances of different regret categories and their corresponding domains, as discussed in Section 1.

**Regret Labelling:** Inspired by (Balouchzahi et al., 2023), the objective of this subtask involves the categorization of Hindi texts into three groups: "Regret by action" (Action), "Regret by inaction" (Inaction), and "No regret". During the annotation process, we classify the datasets into one of two classes: Regret and No Regret. We then further classified regret samples into two sub-classes: "regret by action" and "regret by inaction". The guidelines used for annotating the sub-class categorization of the dataset are stated below, along with sample texts in Table 2

- **No Regret:** There is no indication of regret in the text.
- **Regret by Action (regret of taking action):** Regret by action refers to regretting past decisions or actions, frequently triggered by unfavorable outcomes in either the short or long run.
- **Regret by Inaction (regret about not doing something):** Regret by inaction arises from the absence of a decision or the failure to take action.

**Domain Identification:** Every text in this subtask will be recognized and categorized into one of the following pre-defined domains: (i) Education, (ii) Health, (iii) Career and Finance, (iv) Romance and Relationships, and (v) Other Domains (Balouchzahi et al., 2023).

### 3.3 Annotation Procedure

We gave the annotators detailed guidelines and illustrated different categories of regret. The task of annotating the dataset to identify instances of spoken language. We recruited five annotators with high proficiency in Hindi. The resulting label was decided based on their majority.

- Dataset along with detailed annotation guidelines, were given to five annotators.
- Five annotators underwent one-on-one meetings and interviews to clarify confusion and ensure understanding of the task, discussing labels and guidelines to clarify confusion.
- Finally, after getting the labeled samples from each annotator, the resulting label was decided by the majority of the five annotators.

**Inter-Annotator Agreement:** Inspired by (Falotico and Quatto, 2015),(Balouchzahi et al., 2022), we used Fleiss' kappa as an inter-annotator agreement. It is a common index for evaluating the reliability of agreement among all annotators. For regret detection and domain identification tasks, the strength and reliability of our proposed dataset were 0.84 and 0.87 respectively.

## 4 Methodology

In our proposed dataset, we tested a variety of traditional machine learning and deep learning models to establish benchmarks and evaluate the reliability of the dataset and annotations.

### 4.1 Pre-Processing

This is the most significant part of the prediction model. To pre-process the Hindi text, we utilized the iNLTK library. Unicode characters, HTML/XML elements, URLs, stop words, alphanumeric, and non-alphanumeric values were removed during text preparation. It raises the quality of data and highlights the features.

| Domains | Regret by Inaction | Regret by Action | No Regret |
|---|---|---|---|
| Education | 334 | 277 | 111 |
| Career and Finance | 205 | 326 | 64 |
| Health | 170 | 227 | 36 |
| Romance and Relationships | 905 | 1185 | 608 |
| Other Domains | 138 | 108 | 1481 |

Table 4: Domains subtask distribution of labels

| Dataset | Data_Size | Avg_Sentence_Len | Words | Avg._Word | Char | Vocab |
|---|---|---|---|---|---|---|
| All | 6175 | 42.607 | 263104 | 3.564 | 937937 | 14050 |
| Train | 4940 | 42.890 | 210134 | 3.567 | 749632 | 12470 |
| Test | 1235 | 42.537 | 52970 | 3.554 | 188305 | 5812 |

Table 5: Statistics based on all, Train and Test dataset

## 4.2 Features

We used tf-idf, and bi-gram as feature extraction methods in conventional models, and the pre-trained BERT embeddings were used for deep learning models. Bi-gram word vectorization provides valuable contextual information. Its capability to handle negations, ambiguity, and improves the overall performance of models by providing a more detailed and accurate representation of language and emotional expressions. Our model utilized both tf-idf and bi-gram vectorization features, ensuring a comprehensive representation of document content by considering word frequency importance and sequential relationships across various document levels.

## 4.3 Algorithm

### 4.3.1 Traditional machine learning models

We evaluated our dataset using six traditional machine learning classifiers, namely: (i) Logistic Regression (LR), (ii) Support Vector Machine with Radial Basis Function (RBF), (iii) Multinomial Naive Bayes (MNB), (iv) Random Forest (RF), (v) XGBoost, and (vi) AdaBoost. The classifiers are trained using default parameters presented in Table 6.

### 4.3.2 Deep learning models:

To improve the representation of the Hindi regret text and extract more relevant features, seven deep learning models (i) Long Short-Term Memory (LSTM), (ii) Bi-directional Long Short-Term Memory (Bi-LSTM), (iii) Convolution Neural Network (CNN), (iv) BERT,(v) BERT+LSTM, (vi) BERT+Bi-LSTM and (vii) BERT+CNN were trained and validated.

**CNN architecture:** We implemented a CNN architecture for text classification, consisting of a single convolutional and max-pooling layer succeeded by a dense layer. The model incorporated 128 filters, each with a size of $3 \times 3$ for convolution, coupled with a $5 \times 5$ window for pooling operations. The CNN was configured with categorical cross-entropy loss, used the adam optimizer, applied a softmax activation function for the output layer, and implemented a dropout rate of 0.2 for regularization. The word embedding was set to a dimension of 300, and the model underwent training for 100 epochs. The network accepts embedded vectors of the statements as input and outputs the phrases “Regret by Action”, ”Regret by Inaction “ and “No regret” along with the identification of the domain to which it is related. The hyperparameters of the model are presented in Table 7.

**LSTM architecture:** Each memory cell within an LSTM network is interconnected.

| Estimator | Hyperparameters |
|---|---|
| LR | penalty='l2', C=1.0, solver='lbfgs', max_iter=100, verbose=0, |
| SVM | C=1.0, kernel='rbf,linear',gamma=1.0, cache_size=200, verbose=False, max_iter=-1, random_state=None, |
| MNB | alpha = 0.5 , class_prior = None, fit_prior = True, |
| RF | n_estimators=100, max_depth=10, min_samples_split=2,loss='deviance', |
| XGBoost | lr=0.1, n_est=100, min_samples_split=2, min_sample_leaf=1, max_depth=3, |
| AdaBoost | lr=0.1,est= 200, random_state = None |

Table 6: Important parameters for Machine Learning models

| Model | Parameters |
|---|---|
| LSTM | loss='categorical_crossentropy', num_epoches =100,opti='adam', dropout = 0.2, lr=0.001 |
| Bi-LSTM | loss='categorical_crossentropy', opti='adam', activation='softmax,dropout= 0.2, embedding size=300,epochs =100 |
| CNN | loss='categorical_crossentropy', opti='adam', filters_size=3, activation='softmax, dropout= 0.2,embedding size=300,epochs = 100 |
| BERT | num_words=50000, maxlen=128, pretrained_BERT_mdl='bert-base-uncased' |
| BERT-LSTM | maxlen=128, batch_size=64, model_name = 'bert-base-multilingual-uncased', lr=0.001,epochs=30 ,activation='softmax' |
| BERT-Bi-LSTM | maxlen=128, batch_size=64, pretrained_BERT_model=bert-base-multilingual-uncased, lr=0.001,epochs=30,activation='softmax',embedding_size = 300,epochs = 30 |
| BERT-CNN | loss='categorical_crossentropy', opti='adam', filters_size=3, activation='softmax, pretrained_BERT_model='bert-base-multilingual-uncased,lr=0.001, embedding_size = 300,epochs = 30 |

Table 7: Important parameters of deep models

Furthermore, each memory cell comprises an input gate, a forget gate, an output gate, and a cell state. These gates, in conjunction with the cell state, enable LSTM to effectively preserve extended dependencies in data. Depending on their significance, the gates facilitate the addition or removal of information from the cell state, allowing for nuanced encoding. For a comprehensive understanding of how an LSTM cell functions, you can refer to the detailed description provided in (Hochreiter and Schmidhuber, 1997). Every input sentence is adjusted to be either 64 words in length through padding, or it is shortened to 64 words if the original input exceeds that length. The model's hyperparameters are detailed in Table 7. In contrast to traditional LSTM networks that operate solely in one direction, BiLSTMs (Liu and Guo, 2019) can examine the input sequence in both forward and backward directions. This unique characteristic enables them to capture contextual information from the complete sequence, which can be valuable for text classification, as it allows the model to take the entire input text when making predictions.

The primary distinction between CNNs and Bi-LSTM versions for text categorization is the type of information they can extract from the input data (Balouchzahi et al., 2023). CNNs can automatically learn features from the input text, but the Bi-LSTM can consider contextual information from the complete input sequence.

**BERT:** BERT is a language representation model, which stands for Bidirectional Encoder Representations from Transformers (Devlin et al., 2018). BERT is an attention mechanism that learns the contextual links between words (or subwords) in a text. In its basic setup, the transformer has two separate processes: an encoder that reads the text input and a decoder that provides a task prediction. In contrast to directional models, which read the text input sequentially (from right to left or left to right), the transformer encoder reads the full string of words at once. This feature allows the model to recognize the context of a word based on its surroundings (to the left and right of the word).

**BERT-Deep Learning Models:** From the

above different models, BERT proves to be a powerful model for regret detection in Hindi text. We used three variations of BERT-based models named BERT-LSTM, BERT-Bi-LSTM, and BERT-CNN. All models share common configurations, including a maximum sequence length of 128, a batch size of 64, the utilization of the 'bert-base-multilingual-uncased' pre-trained BERT model, a learning rate of 0.001, the Adam optimizer, a training duration of 30 epochs and applied a softmax activation function for the output layer. BERT-LSTM introduces a specific model name parameter, while BERT-CNN incorporates an additional number of filters and a filter size of 3. Comparing the performance of BERT with other deep learning models, particularly in tasks like regret detection and domain identification, can be interesting. Here the details of embedding algorithms are given below in Table 7.

### 4.4 Evaluation

Precision, Recall, and F1-scores (micro, macro, and weighted averaging methods) are commonly employed metrics and techniques for assessing the performance of classification models. The distinction between these averaging methods lies in how they account for the influence of classes or samples when calculating the final averaged score.

1. Micro-averaged method: A balanced data -set is preferred by the micro-averaged method, which considers the percentage of correctly identified samples among all samples (i.e., each sample contributes equally to the final averaged score).

2. Macro-averaged method: This macro average method, unlike the micro-averaged method, provides an advantage in handling imbalanced datasets by treating all classes equally, irrespective of sample counts. The impact on the final score is greater for classes with more samples.

3. Weighted-averaged method: On the other hand, The weighted-averaged method computes scores by averaging individual scores for each class, considering the size of each class, measured by the number of samples.

## 5 Results and analysis

### 5.1 Regret Detection:

The performance of the conventional machine-learning model for the task of regret detection using tf-idf and a combination of tf-idf and bi-grams is presented in Table 8. The results using tf-idf+bi-gram features show that adding more features improves the performance of these models. SVC with an averaged-macro F1-score of 0.73 using tf-idf + bi-gram features outperformed other ML baseline models for the regret detection task in Hindi text.

The result report of different deep-learning models is listed in Table 9. The result of deep learning models, excluding BERT+CNN is not better compared to machine learning models, possibly due to the absence of fine-tuning in the proposed neural network models. Section 6 discusses the various explanations behind this. The F1-score of BERT-CNN is 0.78. When compared to other machine learning and deep neural network-based models in use, the BERT-CNN model outperforms them in terms of accuracy, recall, and f1-score.

#### 5.1.1 Regret Domain Identification

Tables 10 and 11 display the results of both machine learning baselines and deep learning models in terms of their performance for the domain identification task. Unlike the regret detection subtask, the findings indicate that tf-idf slightly improved the performance of the machine learning models as compared to tf-idf + bi-gram. LR classifier with tf-idf achieved an averaged-macro F1-score of 0.55 as the best-performing ML classifier for the task of domain identification. Furthermore, the results show that the deep learning model BERT+CNN again outperformed those with an averaged-macro F1-score of 0.69.

### 5.2 Analysis

Table 12 displays the comparison between best-performing models for ML and DL models. The results show an improvement in performance using deep learning models with word embedding. For regret detection tasks machine learning models also perform well as compared to deep learning models without a word embedding. Furthermore, based on overall analysis, deep learning models that employ word embedding, specifically BERT, often outperformed conventional machine learning classifiers with tf-idf + bi-grams which could be due to various reasons.

| Regret Detection | | | | | | | | |
|---|---|---|---|---|---|---|---|---|
| **Features** | **Model** | ***Avg. Weighted Scores*** | | | ***Avg. Macro Scores*** | | | **Accuracy** |
| | | Precision | Recall | F1-score | Precision | Recall | F1-score | |
| **tf-idf** | MNB | 0.71 | 0.69 | 0.69 | 0.70 | 0.68 | 0.68 | 0.69 |
| | **SVM** | **0.74** | **0.74** | **0.73** | **0.73** | **0.73** | **0.72** | **0.74** |
| | RF | 0.70 | 0.71 | 0.69 | 0.70 | 0.69 | 0.68 | 0.71 |
| | LR | 0.71 | 0.72 | 0.71 | 0.70 | 0.71 | 0.70 | 0.72 |
| | XGB | 0.72 | 0.73 | 0.72 | 0.71 | 0.71 | 0.71 | 0.73 |
| | AdaBoost | 0.69 | 0.70 | 0.70 | 0.68 | 0.69 | 0.68 | 0.70 |
| **tf-idf + bi-gram** | MNB | 0.73 | 0.72 | 0.72 | 0.72 | 0.71 | 0.71 | 0.72 |
| | **SVM** | **0.75** | **0.75** | **0.74** | **0.74** | **0.74** | **0.73** | **0.75** |
| | RF | 0.68 | 0.68 | 0.64 | 0.68 | 0.66 | 0.64 | 0.68 |
| | LR | 0.73 | 0.73 | 0.73 | 0.72 | 0.72 | 0.72 | 0.73 |
| | XGB | 0.74 | 0.73 | 0.74 | 0.73 | 0.73 | 0.72 | 0.74 |
| | AdaBoost | 0.70 | 0.71 | 0.70 | 0.69 | 0.69 | 0.69 | 0.71 |

Table 8: Results of conventional machine-learning model and features for Regret Detection

| Regret Detection With or Without Embeddings | | | | | | | | |
|---|---|---|---|---|---|---|---|---|
| **Word Embedding** | **Model** | ***Avg. Weighted Scores*** | | | ***Avg. Macro Scores*** | | | **Accuracy** |
| | | Precision | Recall | F1-score | Precision | Recall | F1-score | |
| | LSTM | 0.45 | 0.45 | 0.45 | 0.44 | 0.44 | 0.44 | 0.45 |
| | Bi-LSTM | 0.43 | 0.44 | 0.43 | 0.43 | 0.43 | 0.43 | 0.44 |
| | CNN | 0.44 | 0.54 | 0.46 | 0.41 | 0.52 | 0.44 | 0.54 |
| | **BERT** | **0.71** | **0.68** | **0.67** | **0.70** | **0.67** | **0.66** | **0.68** |
| **BERT** | LSTM | 0.70 | 0.71 | 0.70 | 0.69 | 0.69 | 0.69 | 0.71 |
| | Bi-LSTM | 0.71 | 0.71 | 0.71 | 0.70 | 0.70 | 0.70 | 0.71 |
| | **CCN** | **0.80** | **0.79** | **0.79** | **0.79** | **0.78** | **0.78** | **0.79** |

Table 9: Results of Deep-learning model for Regret Detection

| Regret Domain Identification | | | | | | | | |
|---|---|---|---|---|---|---|---|---|
| **Features** | **Model** | ***Avg. Weighted Scores*** | | | ***Avg. Macro Scores*** | | | **Accuracy** |
| | | Precision | Recall | F1-score | Precision | Recall | F1-score | |
| **tf-idf** | MNB | 0.63 | 0.59 | 0.56 | 0.68 | 0.42 | 0.46 | 0.59 |
| | SVM | 0.66 | 0.64 | 0.62 | 0.71 | 0.48 | 0.52 | 0.64 |
| | RF | 0.59 | 0.58 | 0.51 | 0.58 | 0.34 | 0.32 | 0.58 |
| | **LR** | **0.66** | **0.65** | **0.63** | **0.69** | **0.51** | **0.55** | **0.65** |
| | XGB | 0.62 | 0.62 | 0.60 | 0.62 | 0.47 | 0.50 | 0.62 |
| | AdaBoost | 0.62 | 0.58 | 0.53 | 0.65 | 0.37 | 0.38 | 0.58 |
| **tf-idf + bi-gram** | **MNB** | **0.66** | **0.64** | **0.62** | **0.71** | **0.50** | **0.54** | **0.64** |
| | SVM | 0.65 | 0.63 | 0.61 | 0.68 | 0.49 | 0.52 | 0.63 |
| | RF | 0.61 | 0.58 | 0.50 | 0.61 | 0.33 | 0.30 | 0.58 |
| | LR | 0.66 | 0.64 | 0.61 | 0.72 | 0.48 | 0.52 | 0.64 |
| | XGB | 0.64 | 0.63 | 0.61 | 0.65 | 0.48 | 0.52 | 0.63 |
| | AdaBoost | 0.65 | 0.59 | 0.54 | 0.75 | 0.38 | 0.40 | 0.59 |

Table 10: Results of conventional machine-learning model and features for Regret Domain Identification

Word embedding provides a more in-depth and accurate representation of word meanings and contexts, empowering deep learning models to make more accurate predictions and classifications based on text data. Here in this study, we use BERT as word embedding with deep learning models. The major element within a BERT network is the transformer, an attention mechanism that acquires an understanding of the contextual connections among words (or subwords) in a given text. The transformer consists of two separate procedures: an encoder, which processes the input text, and a decoder, responsible for generating predictions for a given task. Unlike sequential models that read text input sequentially (either from right to left or left to right), the transformer encoder simultaneously processes the entire string of words.

| Regret Domain Identification With or Without Embeddings | | | | | | | | |
|---|---|---|---|---|---|---|---|---|
| **Word Embedding** | **Model** | ***Avg. Weighted Scores*** | | | ***Avg. Macro Scores*** | | | **Accuracy** |
| | | Precision | Recall | F1-score | Precision | Recall | F1-score | |
| | LSTM | 0.37 | 0.40 | 0.38 | 0.26 | 0.26 | 0.25 | 0.40 |
| | Bi-LSTM | 0.37 | 0.41 | 0.33 | 0.28 | 0.27 | 0.26 | 0.41 |
| | CNN | 0.19 | 0.44 | 0.27 | 0.16 | 0.20 | 0.16 | 0.44 |
| | **BERT** | **0.66** | **0.65** | **0.63** | **0.69** | **0.51** | **0.55** | **0.65** |
| **BERT** | LSTM | 0.62 | 0.62 | 0.62 | 0.59 | 0.54 | 0.55 | 0.62 |
| | Bi-LSTM | 0.60 | 0.51 | 0.53 | 0.62 | 0.63 | 0.61 | 0.63 |
| | **CNN** | **0.69** | **0.69** | **0.69** | **0.67** | **0.63** | **0.65** | **0.69** |

Table 11: Results of Deep-learning model for Regret Domain Identification

| Regret Detection | | | | | | | |
|---|---|---|---|---|---|---|---|
| **Model** | ***Avg. Weighted Scores*** | | | ***Avg. Macro Scores*** | | | **Accuracy** |
| | Precision | Recall | F1-score | Precision | Recall | F1-score | |
| SVM (tf-idf+bi-gram) | 0.75 | 0.75 | 0.74 | 0.74 | 0.74 | 0.73 | 0.75 |
| BERT+CNN | 0.80 | 0.79 | 0.79 | 0.79 | 0.78 | 0.78 | 0.79 |
| Regret Domain Identification | | | | | | | |
| LR(tf-idf) | 0.66 | 0.65 | 0.63 | 0.69 | 0.51 | 0.55 | 0.65 |
| BERT+CNN | 0.69 | 0.69 | 0.69 | 0.67 | 0.63 | 0.65 | 0.69 |

Table 12: Best performing models in each learning approach

The BERT model was trained on text from various sources in 104 languages. This multilingual training approach allows BERT to capture general language patterns and semantics across a wide range of languages, making it versatile for tasks involving different linguistic contexts. For the Hindi dataset machine learning algorithms that rely on tf-idf and bi-gram outperform deep learning algorithms without fine-tuning. However, Deep-learning models with word embeddings can enhance their performance by learning and capturing complex text patterns, outperforming traditional machine-learning classifiers.

In general, it is evident from observations that BERT, with its pre-trained, fine-tunable, and detailed word vectors is a better choice for the Hindi dataset when it comes to tasks such as regret detection and domain identification, in contrast to traditional models.

## 6 Error Analysis

We check the model's robustness by manually reviewing a subset of data samples along with their actual labels and the model's predictions, as shown in Table 13.

Classifying regret as regret by action or regret by inaction can be challenging due to their similar expressions and the difficulty in distinguishing them based on language. For example,"एक मौके को चूकना एक पूर्व घटना है। उसके लिए आज भी खुद को कोसते हैं "पछतावा" है!"(Missing an opportunity is a prior event. For that, even today i curse myself for having "regrets"). Additionally, the context in which regret is expressed can influence its classification as a result of action or inaction. For example, a statement like the above "एक मौके को चूकना"((Missing an opportunity) could be interpreted as regret by inaction. However, it could be interpreted as regret by action due to linguistic confusion. Here word "चूकना"(miss) can have multiple meanings like "भूल करना"," गलती करना". Homonyms are words that share the same spelling or pronunciation but have different meanings. This variability in sentence structure can affect the polarity of the text, making linguistic analysis in Hindi more complex. Additionally, the morphological richness of Hindi results in a single root word generat-ing various related words, complicating emotion analysis. Variations in spelling, due to the phonetic nature of the language and diverse dialects, further add to the complexity. Generally, categorizing regret as stemming from an action or from inaction is challenging due to its ambiguous nature.

| Text | Actual Label | Predicted Label |
|---|---|---|
| मुझे सबसे गहरा अफसोस इस बात का है कि दूसरे लोगों को मैं नियंत्रित करने दे रहा हूं कि मैं कौन हूं। | Inaction | Action |
| बहुत कठिन अध्ययन करना वास्तव में कागजात का अध्ययन करने और अभ्यास करने में इतना समय बिताने जैसा है क्योंकि परिवार ट्यूशन का खर्च नहीं उठा सकता है इसलिए मैंने सोचा कि सबसे अच्छा तरीका कम से कम औसत ग्रेड से ऊपर प्राप्त करना है ताकि इससे विश्वविद्यालय में रखा जा सके। मुझे लगता है कि अफसोस इस बात का है कि मुझे पर्याप्त "मज़ा" नहीं मिला, | Inaction | Action |
| मुझे मेरी कायरता पर अफ़सोस है। मैं कुछ भी शुरू करने से पहले ही मेरे अंदर बहुत सारी नेगेटिव सोच आने लगती हैं, जिसके कारण मैं कोई भी नया काम शुरू नहीं कर पाती। मैं आज तक अपनी इसी समस्या से परेशान रहती हूँ। | Inaction | Action |
| एक मौके को चूकना एक पूर्व घटना है। उसके लिए आज भी खुद को कोसते हैं "पछतावा" है! | Inaction | Action |
| मेरा बड़ा पछतावा यह था (और अब भी है) कि मैं दूसरों को अपने भावनाओं के बारे में नहीं बताने का निर्णय नहीं लिया। | Action | Inaction |
| मेरा सबसे बड़ा पछतावा मेरे उस "दोस्त" को गाली न देना और थप्पड़ न मारना है जिसने मेरे अन्य दोस्तों का यौन उत्पीड़न किया | Inaction | Action |

Table 13: Misclassified samples with actual and predicted labels

### 6.1 Errors of inaction class

Originally the text was classified as regret resulting from inaction, but our model predicted it as regret by action. Here inaction is confused with action while describing the inactions individuals took in their past situations, they faced challenges in making accurate decisions and executing actions to resolve the circumstances effectively and timely. In such cases, representing regret stemming from inaction becomes confused with regret from action, due to the presence of several action words in the statement. for example,"मेरा सबसे बड़ा पछतावा मेरे उस दोस्त को गाली न देना और थप्पड न मारना है जिसने मेरे अन्य दोस्तों का यौन उत्पीड़न किया"(My biggest regret is not abusing and slapping my friend who sexually harassed my other friends). It was observed that this statement might simultaneously express regret by inaction and action. If we just discard the negation word "न" from this statement.

### 6.2 Errors of action class

Originally the text was classified as regret resulting from action, but our model predicted it as regret by inaction.

Action is confused with inaction, while describing the actions, individuals committed in their past they faced challenges in making accurate decisions and executing actions to resolve the circumstances effectively and timely. Such as,"मेरा बड़ा पछतावा यह था (और अब भी है) कि मैं दूसरों को अपने भावनाओं के बारे में नहीं बताने का निर्णय नहीं लिया।"(My biggest regret was (and still is) that I didn't decide to tell others about my feelings). In this statement, the way of representing or writing the regret is most confusing and double negatives. Instead of writing "मेरा बड़ा पछतावा यह था (और अब भी है) कि मैं दूसरों को अपने भावनाओं के बारे में बताने का निर्णय लिया।"(My My biggest regret was (and still is) that i decided to tell about my feelings to others). As per the statement, it is clear that people on social platforms are free to express their emotions without being constrained by grammar.

### 6.3 Errors of no-regret class

The text was labeled as no-regret, but a few of them were misclassified by our model predicted it as regret by inaction or regret by action. In case of no regret, most samples are correctly classified as given in the above confusion matrix.

Overall, From the given confusion matrix Figure (1, 2) it is clear that most misclassifications happen in between action and inaction action classes because of their similar expression and context-dependent nature.

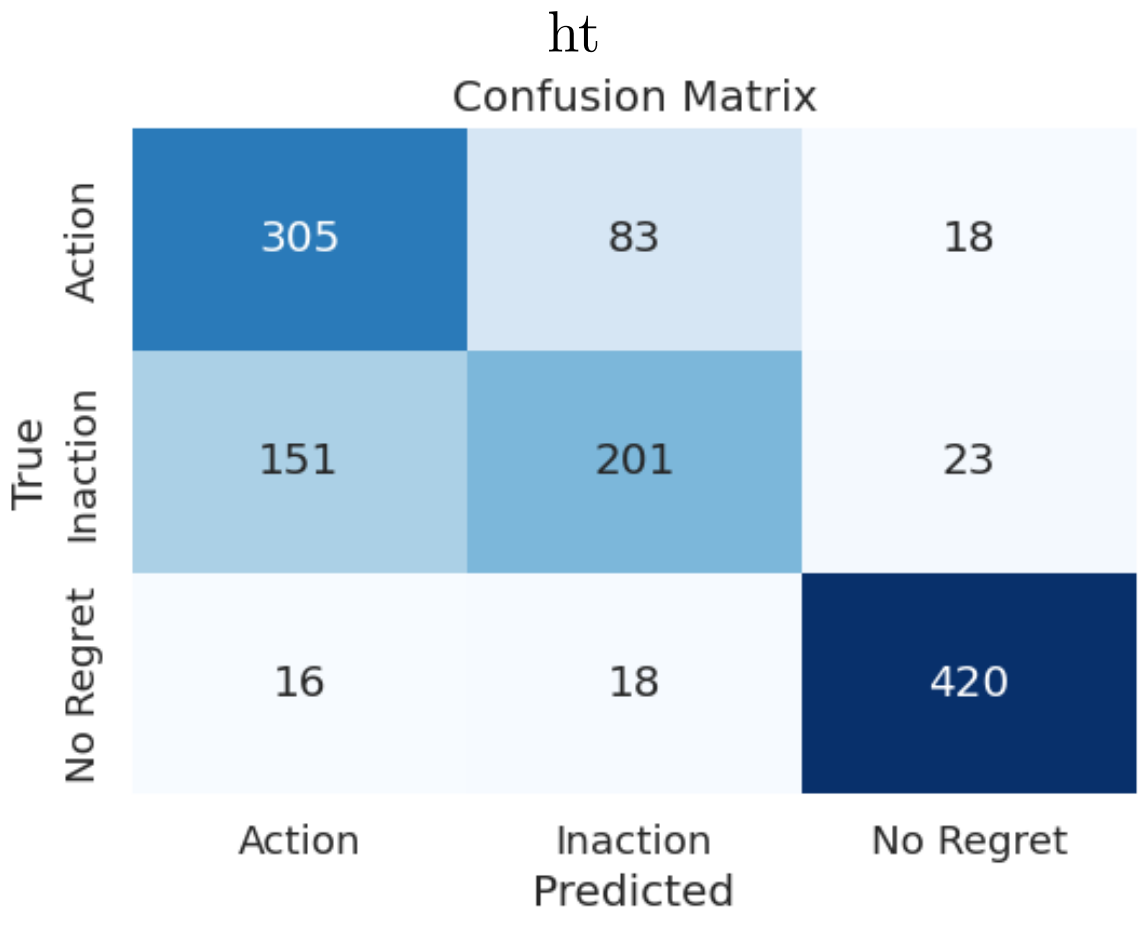

Figure 1: Confusion matrix for SVM model

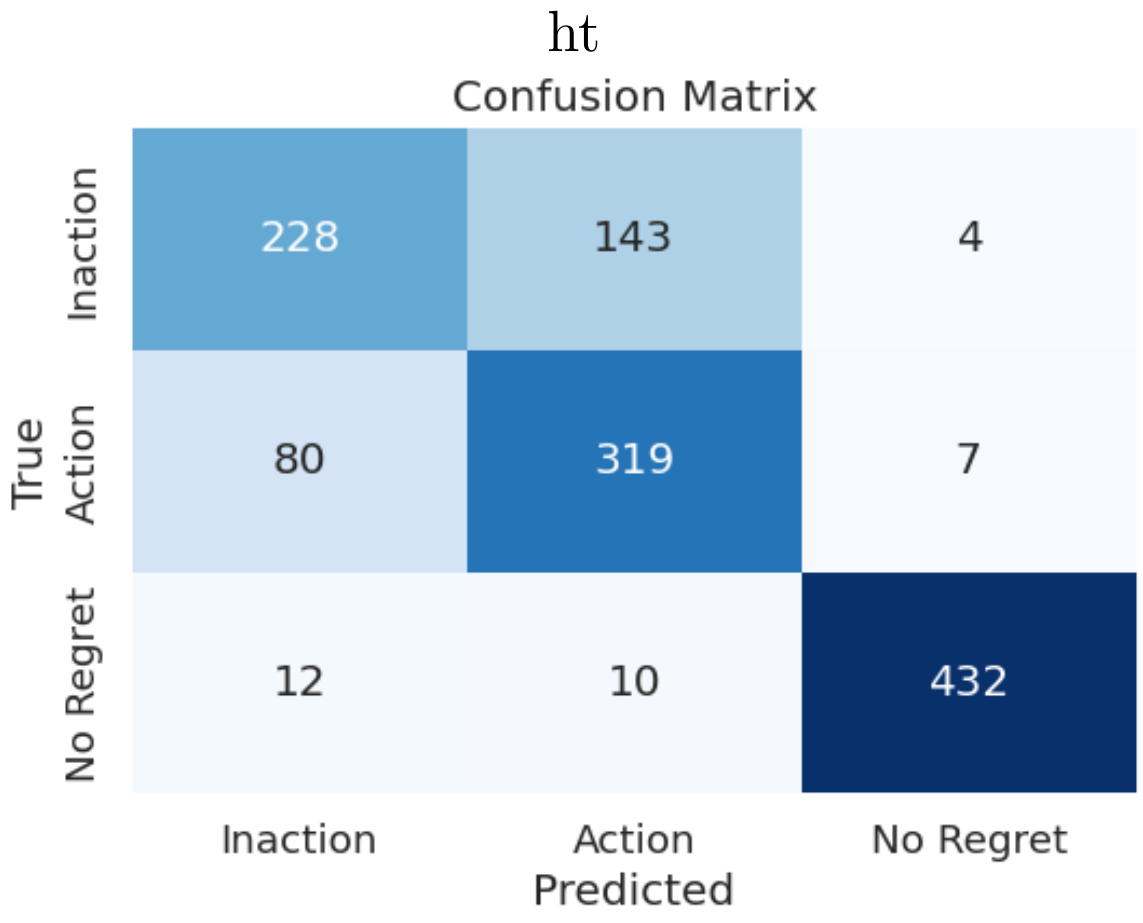

Figure 2: Confusion matrix for CNN model

## 7 Conclusion

Regret is a significant and common emotion that can profoundly affect one's mental health and general well-being. By understanding how individuals express regret on social media, researchers gain insights into the factors influencing regret, its impact on individuals, and potential strategies for managing it. In this article, we provide a detailed description of the Hindi dataset from different sources (Online survey form, YouTube, and Quora), for the purpose of regret detection and domain identification. We have described our dataset development process and annotation guidelines. The dataset statistics indicate that social media users are more likely to express regret about past actions and inaction with an average sentence length of 42.607 in the domain of relationships. In order to benchmark the dataset, we also experimented with different traditional machine learning baselines and deep learning models with stylometric-based features, word bi-grams, tf-idf, and pre-trained BERT embedding for regret detection tasks in the Hindi dataset. In summary, SVM with tf-idf and bi-gram outperformed all machine learning models with an average weighted f1 Score of 0.74%, average macro f1 Score of 0.73%, and accuracy of 0.75% in regret detection task and BERT embedding with CNN model outperformed other models in all experiments with an average weighted f1 Score of 0.79%, average macro f1 Score of 0.78%, and accuracy of 0.79% in regret detection and domain identification task.

## 8 Future Work

In the future, we would redesign the available multi-label/multi-class Hindi Devanagari datasets to incorporate Regretful instances. We would also like to design experiments using large language models (LLM) to evaluate the reasoning and prediction capabilities of the emerging open-source LLMs in differentiating regret from closely related emotions.